\documentclass{amia}
\usepackage{graphicx}
\usepackage[labelfont=bf]{caption}
\usepackage{amsfonts, amsmath, amsthm, amssymb, bbm}
\usepackage[superscript,nomove]{cite}
\usepackage{color}
\usepackage{amsmath}
\usepackage{amssymb}
\usepackage{multirow}

\begin{document}

\title{Avoiding Biased Clinical Machine Learning Model Performance Estimates in the Presence of Label Selection}

\author{Conor K. Corbin$^{1, 3}$, Michael Baiocchi$^{2}$, Jonathan H. Chen, MD, PhD$^{3, 4, 5}$}

\institutes{
    $^1$Department of Biomedical Data Science, Stanford, California, USA \\
    $^2$ Department of Epidemiology and Population Health, Stanford, California, USA \\
    $^3$Center for Biomedical Informatics Research, Stanford, California, USA \\
    $^4$Division of Hospital Medicine, Stanford, California, USA \\
    $^5$ Clinical Excellence Research Center, Stanford, California, USA
}

\maketitle

\textit{When evaluating the performance of clinical machine learning models, one must consider the deployment population.  When the population of patients with observed labels is only a subset of the deployment population (label selection), standard model performance estimates on the observed population may be misleading. In this study we describe three classes of label selection and simulate five causally distinct scenarios to assess how particular selection mechanisms bias a suite of commonly reported binary machine learning model performance metrics.  Simulations reveal that when selection is affected by observed features, naive estimates of model discrimination may be misleading.  When selection is affected by labels, naive estimates of calibration fail to reflect reality. We borrow traditional weighting estimators from causal inference literature and find that when selection probabilities are properly specified, they recover full population estimates. We then tackle the real-world task of monitoring the performance of deployed machine learning models whose interactions with clinicians feed-back and affect the selection mechanism of the labels.  We train three machine learning models to flag low-yield laboratory diagnostics, and simulate their intended consequence of reducing wasteful laboratory utilization.  We find that naive estimates of AUROC on the observed population undershoot actual performance by up to 20\%.  Such a disparity could be large enough to lead to the wrongful termination of a successful clinical decision support tool.  We propose an altered deployment procedure, one that combines injected randomization with traditional weighted estimates, and find it recovers true model performance.}

\section*{Introduction}
Increased access to real-world data has accelerated the use of machine learning (ML) models for precision medicine.  ML practitioners tasked with evaluating the performance of these models must consider the population of patients on which inference will be performed — the target population. If a held out test set used to evaluate a clinical ML model's performance represents only a subset of examples in the target population, model performance metrics estimated using the test set may differ from the model's performance in production.  

One mechanism that drives these situations is label selection (censoring) \cite{leung1997censoring, wang2019machine}.  Consider the time-to-event problem framed as a binary task of predicting all cause 1-year mortality using electronic medical record (EMR) data \cite{avati2018improving}. Without linking to external sources, for example the social security death index, class labels for patients (examples) that lack encounters one year after prediction time may not be observed. An ML practitioner tasked with evaluating this model may find it convenient to only include examples with observed labels in their analysis.  However in deployment, this model will necessarily perform inference on examples whose labels go unobserved. If the model performs preferentially for examples more likely to be observed (a plausible scenario given the model will see more of this data during training) then estimates of model performance on the test set will overshoot the model's real-world performance in deployment.  Label selection is not a phenomena strictly secluded to time-to-event tasks. Consider the ML task of flagging low-yield laboratory test results to mitigate wasteful laboratory utilization; and conversely the task of suggesting additional testing when results are less certain \cite{xu2019prevalence,luo2016using, lidbury2015assessment, chen2016orderrex}.  If a test set only includes the subset of patients for whom laboratory tests were ordered (examples with observed labels) yet the model is deployed on a more general population, performance estimates on the test set may fail to capture true performance in production.

Label selection can occur after successful model development and retrospective validation. Deployed clinical prediction models exhibit feed-back loops and affect prospectively collected data \cite{lenert2019prognostic}.  One way in which a deployed model can induce feed-back is through label selection.  Consider a laboratory prediction task similar to the above designed to alert clinicians of laboratory orders with highly predictable results — this time with the sole purpose of helping clinicians avoid ordering wasteful diagnostics.  Once deployed, and assuming some adherence, labels of laboratory orders the model was most sure of will be go unobserved, as the tests will not be ordered.  Without taking the selection mechanism into consideration, an ML practitioner tasked with monitoring the performance of an already deployed model might incorrectly conclude that it has experienced a degradation in predictive acuity, and perhaps needlessly remove it from production. In reality what the ML practitioner is noticing is not a degradation in performance but rather the model's intended effect — an enrichment of ambiguous cases and removal of those in which test results are mostly certain.  

Though selection mechanisms and their implications on estimates of average treatment effects have been widely studied in causal inference literature \cite{frandsen2015treatment,wu1989estimation,ozenne2020estimation}, label selection in the context of evaluating and monitoring the performance of clinical ML models has received less attention.  In one related study, Powers et al. discuss the evaluation of clinical ML models when the distribution of features in a source population varies from that of a target population \cite{powers2019evaluating}. In ML literature this phenomena is called covariate shift \cite{sugiyama2007covariate, quinonero2008dataset}. Label selection that depends only on observed features (selection at random) is an instance of covariate shift.  We expand upon this work by simulating over several classes of label selection, revealing in which cases ML performance metrics are biased. In another related work, the impact of selection mechanisms on model fitting is assessed \cite{zadrozny2004learning}. While the author does discuss the implications of selection on estimates of model risk, we expand upon their work by addressing its impact on a suite of common ML performance metrics, and discuss when certain kinds of selection mechanisms bias some metrics and not others. Further, we study label selection as it relates to the real-world use case of monitoring the performance of ML models whose deployments feed-back and affect the selection mechanism of prospectively collected data.

In this study we describe three classes of label selection: selection completely at random, selection at random, and selection not at random, borrowing terminology from Little \& Rubin \cite{vink2022roderick}. Adhering to this categorization, we simulate five causally distinct scenarios to assess how different label selection mechanisms impact a suite of commonly tracked binary ML performance metrics, measuring both discrimination and calibration.  In each scenario we assess the ability of traditional weighting estimators from causal inference literature to recover true model performance \cite{hernan2004structural,haneuse2009adjustment}. We then study the effect of label selection on a real-world task.  We train three ML models using EMR data to predict the results of stand-alone laboratory tests, intended to reduce wasteful laboratory utilization.  We evaluate the performance of these models retrospectively, and then simulate their deployment and feed-back on new data (label selection).  We measure inconsistencies between observed and actual performance, and then propose a simple correction method that couples injected randomization into the clinical decision support alerting mechanism with traditional weighting estimators. Simulations reveal that this approach corrects disparities between observed and actual performance using only the observed data, without the need to assume away unmeasured confounding.

\section*{Methods}
\subsection*{Simulation study}
Here we study the effect of label selection on common binary ML model performance metrics.  We discuss three broad categories of selection mechanisms: selection completely at random, selection at random, and selection not at random. We simulate five causally distinct scenarios. In each scenario we estimate model performance in terms of both discrimination and calibration using a suite of commonly tracked performance metrics. Discrimination is tracked with estimates of specificity, sensitivity, accuracy, positive predictive value (PPV), area under the receiver operating characteristics curve (AUROC) and area under the precision/PPV recall curve (AUPRC). Threshold dependent metrics are estimated with the threshold set to $t=0.5$. Calibration is estimated with calibration plots; that is, we bin predictions into $n=5$ equal width buckets and compare the average predicted probability within the bucket to the actual prevalence of the outcome.  We track all performance metrics across the full and observed populations for all scenarios. We then assess the ability of traditional weighting estimators from causal inference literature to recover discrepancies in model performance between full and observed populations. 

\subsubsection*{Data generating process}
In each scenario we sample a dataset $\{x^i, y^i, s^i\}_{i=0}^{n}$ of size $n=10,000$ from a joint distribution $P(X, Y, S)$ that we specify. Each example is a tuple of features vector $x^i$, binary class label $y^i$, and binary selection variable $s^i$ indicating whether the class label of the example is observed.  For all three scenarios, $X$ and $Y$ are sampled according to the following data generating process where $\sigma(x)$ is the sigmoid function. 

\begin{eqnarray}
	&X_1, X_2& \sim Uniform(\alpha, \beta) \\
    &Y& \sim Bernoulli(p=\sigma(\omega_1 x_1 +\omega_2 x_2 + \gamma)) \quad \text{where} \quad \sigma(x) = \frac{1}{1+e^{-x}}
   \end{eqnarray}We allow a hypothetical clinical ML model $h(x) : X\rightarrow Y$ to  perfectly specify the mapping from $X$ to $Y$ by assigning it the following functional form: $h(x) := \sigma(\omega_1 x_1 +\omega_2 x_2 + \gamma)$.  For all simulated scenarios, we set $\alpha, \beta := 2$, $\omega_1, \omega_2 := 1$, and  $\gamma := 0$.  Note that although we have a perfectly specified model, irreducible error induced by stochastic sampling of $Y$ exists leading to imperfect discrimination acuity \cite{james1997generalizations}.  Our model however is perfectly calibrated.  Our data generating process induces a decision boundary on the line described by $\omega_1 X_1 + \omega_2 X_2 + \gamma = 0$. 
   
\subsubsection*{Selection completely at random}
In \textbf{Scenario 1} we simulate an environment where class labels are selected completely at random. Concretely, our selection indicator $S$ is independent of both $X$ and $Y$.  We incorporate this into our data generating process by sampling $S$ from a Bernoulli distribution with a constant parameter set to $\pi_1 := 0.5$. 
\begin{eqnarray}
S_{scenario 1} \sim Bernoulli(p=\pi_1)
\end{eqnarray}
\subsubsection*{Selection at random}
In \textbf{Scenario 2} and \textbf{Scenario 3} we simulate environments where selection is affected by the observed features. In \textbf{Scenario 2} (select hard) selection is more likely on examples difficult to predict. We induce this behavior by sampling $S$ such that it takes a value of 1 more frequently when the example's feature vector resides closer to the model's decision boundary.  We sample $S$ according to the feature vector's distance to the decision boundary, and apply an exponential function such that the probability of observing the data is 1 when the point resides on the decision boundary and trends towards zero as the distance increases.  In \textbf{Scenario 3} (select easy) we simulate the inverse phenomena. We sample $S$ such that it takes the value one with higher probability when the example's feature vector is further from the decision boundary.  We similarly induce this by setting the parameter of the Bernoulli $S$ is sampled from such that the probability of selection is $1$ when the feature vector resides at the edge of the feature space furthest from the decision boundary and trends towards zero as it gets closer.  We make this concrete below.
\begin{eqnarray}
&S_{scenario_2}& \sim Bernoulli(p=exp(\frac{-|\omega_1 x_1 + \omega_2 x_2 + \gamma|}{\sqrt{\omega_1^2 + \omega_2^2}})) \\
&S_{scenario_3}& \sim Bernoulli(p=exp(\frac{|\omega_1 x_1 + \omega_2x_2 + \gamma|}{\sqrt{\omega_1^2 + \omega_2^2}} - \delta)) \quad \text{where} \quad \delta = max(\frac{|\omega_1 x_1 + \omega_2 x_2 + \gamma|}{\sqrt{\omega_1^2 + \omega_2^2}})
\end{eqnarray}
\subsubsection*{Selection not at random}
Here we simulate environments where selection is affected by the label itself. In \textbf{Scenario 4} (select negative) selection is more likely when $y=0$.  In \textbf{Scenario 5} (select positive) selection is more likely when $y = 1$.  We induce this by sampling $S$ according to the distribution defined below. In \textbf{Scenario 4} we set $\pi_1 := 0.5$ and $\pi_2 := 1$. In \textbf{Scenario 5} we set $\pi_1 := 1$ and $\pi_2 := 0.5$. 
\begin{eqnarray}
S_{scenario 4, 5} \sim Bernoulli(p=\pi_1y + \pi_2(1-y))
\end{eqnarray}
\subsubsection*{Weighted model performance estimates}
Across each scenario we track model performance in the full and observed populations using a suite of common binary ML performance metrics, assessing both discrimination and calibration acuity.  We then borrow traditional weighting techniques from causal inference literature to define weighted performance estimates, and assess their ability to recover disparities in observed vs full populations. We construct weighted estimates of each performance metric by assigning weight to each example in the observed population equal to the inverse probability of observing it. Note we know the probability of observing each example because we explicitly define it.  In practice, this probability is instead estimated.  This re-weighing procedure is commonly conducted in causal inference literature to correct for selection bias in average treatment effect estimates, and is known as inverse probability weighting (IPW) \cite{hernan2004structural}. The IPW estimate of an arbitrary performance metric $m(h(x^i), y^i)$ that maps a single example's prediction and label to a score is defined below.  IPW estimates of AUROC and AUPRC (both undefined for a single example) are generated by building up an IPW ROC and PR curve and performing numerical integration.  

\begin{eqnarray}
\widehat{Metric}_{IPW} &=& \frac{1}{W} \sum_{i=1}^N \frac{m(h(x^i), y^i)}{P(s^i = 1)} \quad \text{where} \quad W = \sum_{i=1}^N \frac{1}{P(s^i = 1)} 
\end{eqnarray}

\subsection*{Real world application: monitoring model performance in production}
We now move to the real-world problem of monitoring clinical ML models hypothetically deployed within the EMR's production environment.  Consider the task of developing and deploying ML models that use EMR data to flag low-yield laboratory diagnostics. Many laboratory tests ordered within the hospital are ordered even though their results are highly predictable. Flagging these tests at order time has the potential to cut down on wasteful ordering behavior — saving patients from burdensome blood draws and hospital operational costs.  Here we describe our methods for training, retrospectively evaluating, and simulating the deployment of three clinical ML models tasked with flagging laboratory orders with predictable results. In our simulated deployment, we study the effect of model feed-back loops, and propose a method to recover true performance on the full deployment population.

\subsubsection*{Data source and cohort extraction}
We used the \textbf{STA}nford \textbf{R}esearch \textbf{R}epository (STARR) to extract de-identified patient medical records \cite{datta2020new}. STARR contains electronic health record data collected from over 2.4 million unique patients spanning 2009-2021 who have visited Stanford Hospital (academic medical center in Palo Alto, California), ValleyCare hospital (community hospital in Pleasanton, California) and Stanford University Healthcare Alliance affiliated ambulatory clinics.  Using this repository we extracted cohorts of patients for whom three stand-alone laboratory procedures (Hematocrit, Troponin I, Sodium) had been ordered between the years 2015 and 2020.  Ten thousand laboratory procedures per year were randomly sampled from the total dataset to yield the final cohort. Train, validation, and test sets were constructed by year (training set 2015-2018, validation set 2019, test set 2020).  Prediction time was defined as the timestamp associated with the order of the laboratory procedure.

\subsubsection*{Feature representation}
For each task (laboratory test), a timeline of medical events was constructed from structured electronic medical record data available before prediction time.  A mix of categorical and numerical data elements were included as events in the feature set. Categorical features included diagnosis (ICD 10) codes on a patient's problem list, procedure and medication orders, and demographic variables including race and sex. Numerical features included prior laboratory results and the patient's age at prediction time. Numerical features were discretized into tokens based on the percentile values they assumed in the training set distribution, such that they could be embedded alongside the already token like categorical features.  All diagnosis codes prior to prediction time were included in the patient's constructed timeline. Procedure and medication orders placed within 28 days of prediction time were included, as were laboratory results made available within 14 days of prediction time.  Each timeline was then constructed as a series of days, and features taking place on the same day were grouped together. Demographic variables that lacked association with a particular day were included on the patient's timeline as if they were made available in the final day.

\subsubsection*{Model description, training procedure and evaluation}
Sequences of patient days were embedded into a $512$ dimensional representation, and fed into a gated recurrent unit layer with a hidden representation size of $256$.  The final hidden representation was input into a classification layer parameterized by a feed-forward neural network consisting of two linear layers, ReLU activations, and dropout to output a single score in $\mathbb{R}$. Dropout was set to 0.2 across all tasks.  Models were trained end-to-end with a binary cross entropy loss function using Adam optimization with learning rate set to $10^{-4}$ and weight decay set to $10^{-5}$.  The max number of epochs was set to 200, a learning rate scheduler reduced the learning rate by a power of ten after every 50 epochs, and early stopping was used to terminate model training if validation set AUROC failed to improve after 25 epochs.  The model with the best validation AUROC across epochs was selected for each task, and evaluated on the test set.  We measure each model's discrimination acuity by estimating the ROC curve, and calibration acuity with calibration plots. 

\subsubsection*{Simulating model deployment and feed-back loop}
We simulate the deployment of all three models into the EMR system and their resulting feed-back mechanisms (label selection) using predictions and labels on the test sets. We model a production use case where upon order entry an alert is displayed to clinicians suggesting not to order the laboratory test if the probability of an abnormal lab is above threshold  $p_t$ or below $1 - p_t$, consistent with the hypothesized use case from existing literature \cite{xu2019prevalence}. We track performance in terms of AUROC.

\subsubsection*{Correcting model performance estimates in production under label selection}
Assuming complete adherence to model suggestions, the probability of observing the labels of laboratory tests with predicted probabilities above $p_t$ or below $1-p_t$ is zero. In practice, clinician adherence to model suggestions will not be 100\%.  One could imagine estimating the adherence probability (fraction of alerts accepted) and use a function of its inverse $\frac{1}{1-p_a}$ to weigh examples where alerts were shown.  The problem with this approach however is that clinician adherence would likely depend on patient characteristics, and likely some not recorded in the EMR — leaving our selection probabilities misspecified. Instead we propose injecting randomization into the alert triggering procedure; that is, we propose a production pipeline where an alert will only trigger if 1) the predicted probability of the laboratory test is above $p_t$ or below $1-p_t$, and 2) the result of a single sample from a Bernoulli distribution with parameter $p_{withhold}$ is zero. This procedure ensures that the resulting selection mechanism is at random — only dependent on the output of the model (a function of observed features) and a random number generator.  Further, we know the selection probability of each example because we explicitly define it. In our simulations, we ablate examples in our test set where an alert would hypothetically trigger irrespective of clinician adherence. We track actual, observed, and weighted estimates. Weighted estimates are computed by re-weighing each of the alert-triggering examples by the inverse of the alert withholding probability $\frac{1}{p_{withhold}}$.  Weights for non-alert-triggering examples are set to 1. In our simulations we perform two parameter sweeps.  First we sweep $p_t$ from 0.99 to 0.51 at $p_{withold}$ set to 0.05.  Next we sweep $p_{withold}$ from 0.99 to 0.01 at $p_t$ set to 0.9.  We simulate each parameter setting 1000 times and estimate actual, observed and weighted AUROC, reporting mean values and intervals between the 2.5 and 97.5 percentiles. 


\section*{Results}

\subsection*{Simulation study}
Here we show the results of our simulation study, highlighting how different selection mechanisms bias particular binary ML performance metrics.  We discuss metrics that measure discrimination and calibration separately — reporting full, observed and weighted estimates across all five scenarios.

\subsubsection*{Evaluating model discrimination under label selection}
In Table 1 we we show the how metrics that assess discrimination acuity are impacted by varying selection mechanism. Instances when estimates on the observed population differ from full population estimates are bolded.  In Scenario 1 (selection completely at random)  label selection was independent of both the features and labels. Our simulations reveal no difference between actual, observed, and weighted performance estimates across all six measures, save for a reduction in power due to smaller sample size in the observed population.

In \textbf{Scenario 2} and \textbf{Scenario 3} (selection at random), simulations reveal differences between performance estimates on the full and observed populations.  In \textbf{Scenario 2} (select hard), estimates using only the observed population undershoot actual performance across all six measures.  In \textbf{Scenario 3} (select easy), estimates on the observed population overshoot true performance. Weighted estimates of all performance measures successfully recover true performance using only the observed data. 

In \textbf{Scenario 4} and \textbf{Scenario 5} (selection not at random) simulations reveal variation between observed and actual estimates of PPV and AUPRC, but no variation in estimates of sensitivity, specificity, accuracy, or AUROC.   We note that because accuracy is a weighted average of sensitivity and specificity (weighted by class prevalence) this metric would vary in situations where sensitivity and specificity took on different values.  In \textbf{Scenario 4} (select negative) both PPV and AUPRC estimates in the observed population are lower than estimates on the full population. \textbf{In Scenario 5} (select positive) PPV and AUPRC estimates are higher than in the full population. Note that baseline performance in terms of PPV and AUPRC depend on the prevalence of the positive class, which vary in the observed and full population in both scenarios.   Though estimates of PPV and AUPRC vary between populations, estimates relative to the baseline performance remains stable. In both scenarios, weighted estimates on the observed population recover actual performance on the full population. 

\begin{table}[]
\centering
\small
\caption{Discrimination performance metrics across selection scenarios: Scenario 1 (selection completely at random), Scenario 2 (select hard), Scenario 3 (select easy), Scenario 4 (select negative), Scenario 5 (select positive).}
\begin{tabular}{|l|l|l|l|l|l|l|}
\hline
\textbf{Metric}                       & \textbf{Estimator}         & \textbf{Scenario 1}   & \textbf{Scenario 2}            & \textbf{Scenario 3}            & \textbf{Scenario 4}            & \textbf{Scenario 5}            \\ \hline
\multirow{3}{*}{\textbf{Sensitivity}} & \textit{\textbf{Actual}}   & 0.75 {[}0.74, 0.77{]} & 0.75 {[}0.74, 0.77{]}          & 0.75 {[}0.74, 0.77{]}          & 0.75 {[}0.74, 0.77{]}          & 0.75 {[}0.74, 0.77{]}          \\ \cline{2-7} 
                                      & \textit{\textbf{Observed}} & 0.75 {[}0.74, 0.77{]} & \textbf{0.63 {[}0.6, 0.65{]}}  & \textbf{0.85 {[}0.82, 0.87{]}} & 0.75 {[}0.74, 0.77{]}          & 0.75 {[}0.74, 0.77{]}          \\ \cline{2-7} 
                                      & \textit{\textbf{Weighted}} & 0.75 {[}0.74, 0.77{]} & 0.75 {[}0.72, 0.79{]}          & 0.75 {[}0.72, 0.79{]}          & 0.75 {[}0.74, 0.77{]}          & 0.75 {[}0.74, 0.77{]}          \\ \hline
\multirow{3}{*}{\textbf{Specificity}} & \textit{\textbf{Actual}}   & 0.75 {[}0.74, 0.77{]} & 0.75 {[}0.74, 0.77{]}          & 0.75 {[}0.74, 0.77{]}          & 0.75 {[}0.74, 0.77{]}          & 0.75 {[}0.74, 0.77{]}          \\ \cline{2-7} 
                                      & \textit{\textbf{Observed}} & 0.75 {[}0.74, 0.77{]} & \textbf{0.63 {[}0.6, 0.65{]}}  & \textbf{0.85 {[}0.82, 0.87{]}} & 0.75 {[}0.74, 0.77{]}          & 0.75 {[}0.74, 0.77{]}          \\ \cline{2-7} 
                                      & \textit{\textbf{Weighted}} & 0.75 {[}0.74, 0.77{]} & 0.75 {[}0.72, 0.79{]}          & 0.75 {[}0.72, 0.79{]}          & 0.75 {[}0.74, 0.77{]}          & 0.75 {[}0.74, 0.77{]}          \\ \hline
\multirow{3}{*}{\textbf{PPV}}         & \textit{\textbf{Actual}}   & 0.75 {[}0.74, 0.77{]} & 0.75 {[}0.74, 0.77{]}          & 0.75 {[}0.74, 0.77{]}          & 0.75 {[}0.74, 0.77{]}          & 0.75 {[}0.74, 0.77{]}          \\ \cline{2-7} 
                                      & \textit{\textbf{Observed}} & 0.75 {[}0.74, 0.77{]} & \textbf{0.63 {[}0.6, 0.65{]}}  & \textbf{0.85 {[}0.82, 0.87{]}} & \textbf{0.61 {[}0.59, 0.62{]}} & \textbf{0.86 {[}0.85, 0.87{]}} \\ \cline{2-7} 
                                      & \textit{\textbf{Weighted}} & 0.75 {[}0.74, 0.77{]} & 0.75 {[}0.71, 0.79{]}          & 0.75 {[}0.72, 0.79{]}          & 0.75 {[}0.74, 0.77{]}          & 0.75 {[}0.74, 0.77{]}          \\ \hline
\multirow{3}{*}{\textbf{Accuracy}}    & \textit{\textbf{Actual}}   & 0.75 {[}0.75, 0.76{]} & 0.75 {[}0.75, 0.76{]}          & 0.75 {[}0.75, 0.76{]}          & 0.75 {[}0.75, 0.76{]}          & 0.75 {[}0.75, 0.76{]}          \\ \cline{2-7} 
                                      & \textit{\textbf{Observed}} & 0.75 {[}0.74, 0.77{]} & \textbf{0.63 {[}0.61, 0.64{]}} & \textbf{0.85 {[}0.83, 0.86{]}} & 0.75 {[}0.74, 0.76{]}          & 0.75 {[}0.74, 0.76{]}          \\ \cline{2-7} 
                                      & \textit{\textbf{Weighted}} & 0.75 {[}0.74, 0.77{]} & 0.75 {[}0.73, 0.78{]}          & 0.75 {[}0.73, 0.78{]}          & 0.75 {[}0.74, 0.76{]}          & 0.75 {[}0.74, 0.76{]}          \\ \hline
\multirow{3}{*}{\textbf{AUROC}}       & \textit{\textbf{Actual}}   & 0.84 {[}0.83, 0.84{]} & 0.84 {[}0.83, 0.84{]}          & 0.84 {[}0.83, 0.84{]}          & 0.84 {[}0.83, 0.84{]}          & 0.84 {[}0.83, 0.84{]}          \\ \cline{2-7} 
                                      & \textit{\textbf{Observed}} & 0.84 {[}0.82, 0.85{]} & \textbf{0.68 {[}0.66, 0.7{]}}  & \textbf{0.91 {[}0.9, 0.93{]}}  & 0.84 {[}0.83, 0.84{]}          & 0.84 {[}0.83, 0.84{]}          \\ \cline{2-7} 
                                      & \textit{\textbf{Weighted}} & 0.84 {[}0.82, 0.85{]} & 0.84 {[}0.8, 0.87{]}           & 0.84 {[}0.82, 0.85{]}          & 0.84 {[}0.83, 0.84{]}          & 0.84 {[}0.83, 0.84{]}          \\ \hline
\multirow{3}{*}{\textbf{AUPRC}}       & \textit{\textbf{Actual}}   & 0.83 {[}0.82, 0.84{]} & 0.83 {[}0.82, 0.84{]}          & 0.83 {[}0.82, 0.84{]}          & 0.83 {[}0.82, 0.84{]}          & 0.83 {[}0.82, 0.84{]}          \\ \cline{2-7} 
                                      & \textit{\textbf{Observed}} & 0.83 {[}0.82, 0.85{]} & \textbf{0.68 {[}0.65, 0.71{]}} & \textbf{0.91 {[}0.89, 0.92{]}} & \textbf{0.73 {[}0.71, 0.74{]}} & \textbf{0.9 {[}0.9, 0.91{]}}   \\ \cline{2-7} 
                                      & \textit{\textbf{Weighted}} & 0.83 {[}0.82, 0.85{]} & 0.83 {[}0.76, 0.88{]}          & 0.83 {[}0.81, 0.85{]}          & 0.83 {[}0.82, 0.84{]}          & 0.83 {[}0.82, 0.84{]}          \\ \hline
\end{tabular}
\end{table}

\subsubsection*{Evaluating model calibration under label selection}
In Figure 1 we show model calibration estimates in the form of calibration plots. Because our model is perfectly specified, it is perfectly calibrated. In Figure 1 we omit the calibration curve on the full population, and focus on calibration estimates on the observed population (both with and without weighting). In \textbf{Scenario 1} (selection completely at random) calibration estimates remain stable across populations estimates, save for a reduction in power due to smaller sample size in the observed population.

In \textbf{Scenario 2} and \textbf{Scenario 3} (selection at random) simulations also reveal no change in calibration estimates across populations, except for reductions in power. Distinct from \textbf{Scenario 1},  power is reduced differentially across each of the binned calibration estimates. In \textbf{Scenario 2} (select hard), we see a reduction in power in bins where the model is most confident (further from 0.5 predicted probability). In Scenario 3 (select easy), power is reduced in binned calibration estimates where the predicted probability is closer to 0.5. 

In \textbf{Scenario 4} and \textbf{Scenario 5} (selection not at random), calibration estimates differ when computed on the observed and full populations. In \textbf{Scenario 4} (select negative), the observed calibration plot suggests that the model is over predicting risk even though in reality the model is perfectly calibrated. In \textbf{Scenario 5} (select positive) the observed calibration plot suggests the model is under predicting risk.  In both scenarios, weighted estimates recover the true calibration curve.   

\begin{figure}[t!]
\centering
\includegraphics[scale=0.14]{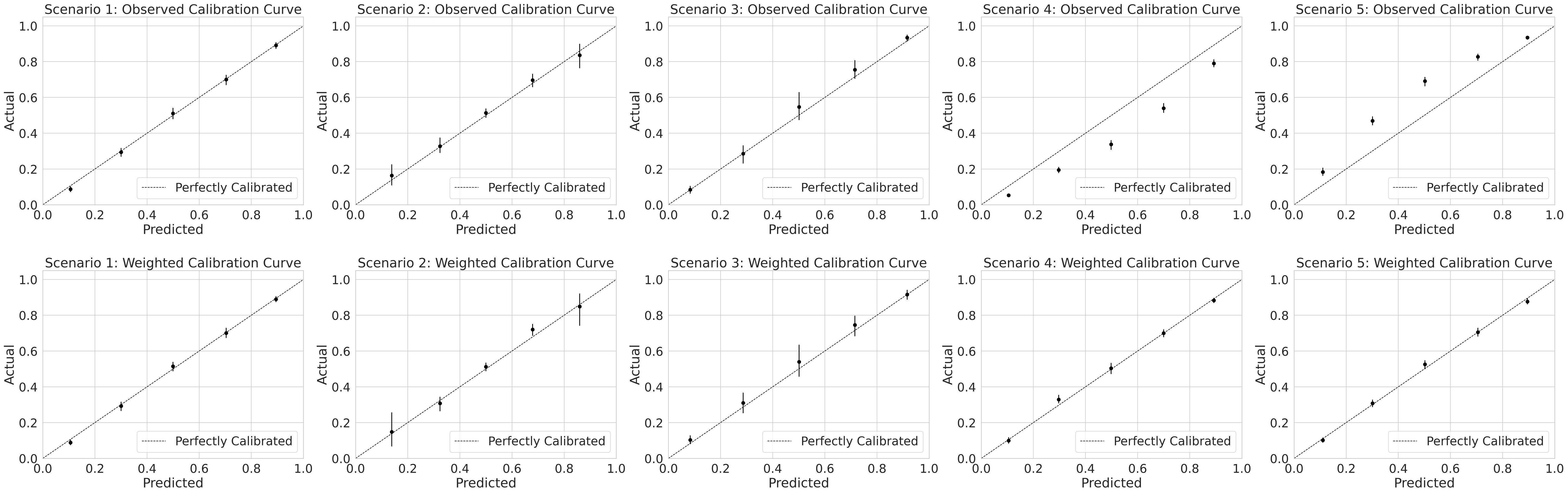}
\caption{Model calibration estimates across selection scenarios}
\label{fig3}
\end{figure}

\subsection*{Real world application: monitoring model performance in production}
Here we present results of training, evaluating, and simulating the deployment of three clinical ML models intended to reduce unnecessary laboratory utilization. We begin this section by presenting model performance in the standard way on a held out retrospective test set.  We then shift to presenting simulated results where model deployment induces label selection, and highlight differences in estimates using observed and actual deployment populations.  We conclude the section by evaluating whether our proposed deployment workflow corrects disparities between observed and full populations.

\subsubsection*{Retrospective evaluation of stand-alone laboratory prediction models}
We trained clinical ML models to predict the results of three stand-alone laboratory procedures: hematorcrit, troponin I and sodium.  Each cohort contained a total of 60,000 randomly sampled laboratory procedures stratified by year between 2015 and 2020.  Training sets included 40,000 laboratory tests between 2015 and 2018.  Validation sets contained 10,000 laboratory tests in ordered in 2019.  Test sets contained 10,000 laboratory tests ordered in 2020.  Prevalence of the outcome in each test set was 0.79 [0.78, 0.80], 0.57 [0.57, 0.58] and 0.27 [0.26, 0.28] for hematocrit, troponin I and sodium tasks respectively.  Performance on the test set for each of the three tasks in terms of both discrimination (ROC curve) and calibration (calibration plots) are shown in Figure 2.  

\begin{figure}[t!]
\centering
\includegraphics[scale=0.21]{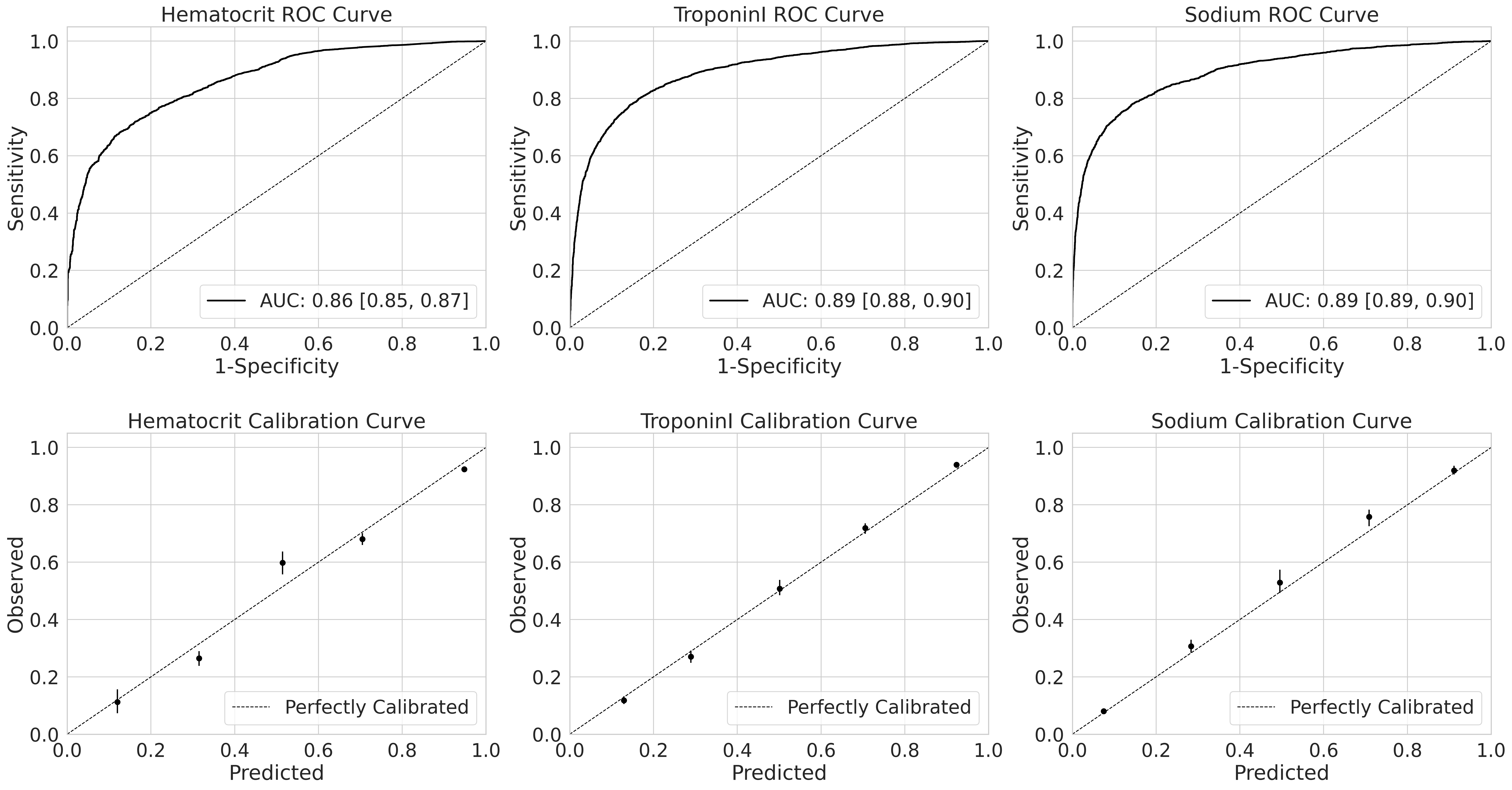}
\caption{Model performance on held out test sets}
\label{fig3}
\end{figure}

\subsubsection*{Simulating model deployment}
In Figure 3 we show the results of our simulated model deployment.  In the top row, we show actual, observed and weighted AUROC estimates over a range of model alert trigger thresholds $p_t$. Recall we trigger an alert upon order entry suggesting not to order the laboratory test if 1) the probability estimate of the model is greater than $p_t$ or below $1-p_t$ and 2) a single sample from a Bernoulli with parameter $p_{withhold}$ is zero. We sweep values of $p_t$ from 0.99 to 0.51. Simulations reveal across all three models that observed performance initially decays, reaches a minima, and then increases towards actual performance as $p_t$ decreases from 0.99 to 0.51. The initial decline in observed performance occurs as observations further from the decision boundary are differentially selected, as in \textbf{Scenario 2} (select hard). After the minima is reached, observed model performance then trends back to actual performance — as the pool of examples failing to trigger an alert become small enough to be outweighed by observations where an alert would have triggered had we not injected random alert withholding probability $p_{withhold}$, here set to 0.05.  Importantly our weighted estimates of AUROC successfully recover actual performance across all values of $p_t$. 

In the second row of Figure 3, we show actual, observed and weighted model performance estimates as a function of our randomly injected alert withholding probability $p_{withhold}$. Here we set $p_t$ (varied in row above) to 0.9. We sweep values of $p_{withhold}$ ranging from 0.99 to 0.01. Our simulations reveal that as $p_{withhold}$ trends toward zero (leading to more alerts and thus more missing labels), AUROC on the observed population decays further from AUROC on the actual deployment population.  Importantly, weighted estimates successfully recover actual performance across all values of $p_{withhold}$. It should be noted that the variance of weighted AUROC estimates increases as $p_{withhold}$ shrinks to zero, which in turn explodes our weights $\frac{1}{p_{withold}}$. This is a known criticism of IPW estimates, where the precision of the estimators decrease with larger weights \cite{hernan2004structural}.

\section*{Discussion}
In this study we assess how different types of label selection yield biased clinical ML model performance estimates.  We simulate five causally distinct scenarios, and probe how they impact a suite of typically reported ML performance metrics.  We categorize our scenarios into three broad classes: 1) selection completely at random, where selection is independent of both features and labels, 2) selection at random, where selection is affected by features, and 3) selection not at random, where selection is affected by the labels themselves.  Our simulations reveal that when selection is completely at random, model performance estimates are unaffected, save for a reductions in power.  When selection is at random, performance metrics that measure a model's discrimination acuity (sensitivity, specificity, positive predictive value, accuracy, AUROC, AUPRC) can be biased if naively estimated on the observed population. If selection occurs such that examples harder to predict are preferentially selected (\textbf{Scenario 2}), discrimination metrics may underestimate true performance. If selection occurs such that easier examples are preferentially selected (\textbf{Scenario 3}), discrimination metrics may overestimate true performance. While selection at random can yield biased estimates of model discrimination acuity, estimates of model calibration on the observed population remain consistent.  Power to estimate calibration in regions of the feature space with more missing labels is reduced.  When selection is only affected by observed features, labels are independent of the selection mechanism when conditioned on features, implying $p(y \mid  x) = p(y \mid x, s=1)$. 

When selection occurs not at random, performance metrics on the observed population that condition on the class label (sensitivity, specificity, AUROC) remain consistent while calibration estimates veer from truth. This is revealed in \textbf{Scenario 4} and \textbf{Scenario 5}, which result in estimates of sensitivity, specificity, and AUROC that do not differ between observed and full populations, while calibration plots indicate over and under estimation of risk. In our simulations we keep our model (which is perfectly specified) fixed. If a model had instead been fit to the observed data, calibration plots may imply well calibrated models when in reality they would be under and over estimating risk respectively.  An important note, discrimination metrics sensitive to the prior $p(y)$ like PPV (precision), AUPRC, and accuracy may not reflect their true population values when selection is not at random. In our simulations, we see a change in PPV and AUPRC.  Accuracy, effectively a weighted average of sensitivity and specificity weighted by class prevalence, remained stable only because in our case sensitivity and specificity were equal at our cutoff probability threshold of 0.5.  In our simulations, traditional weighting estimators borrowed from causal inference literature effectively corrected performance estimates, recovering the full population parameters. These estimators rely on well specified estimates of the selection probabilities.  In our simulations these were known, in practice they must be estimated.  Our simulations reveal how weighted machine learning model performance estimators can recover full population parameters in the presence of missing labels. 

\begin{figure}[t!]
\centering
\includegraphics[scale=0.17]{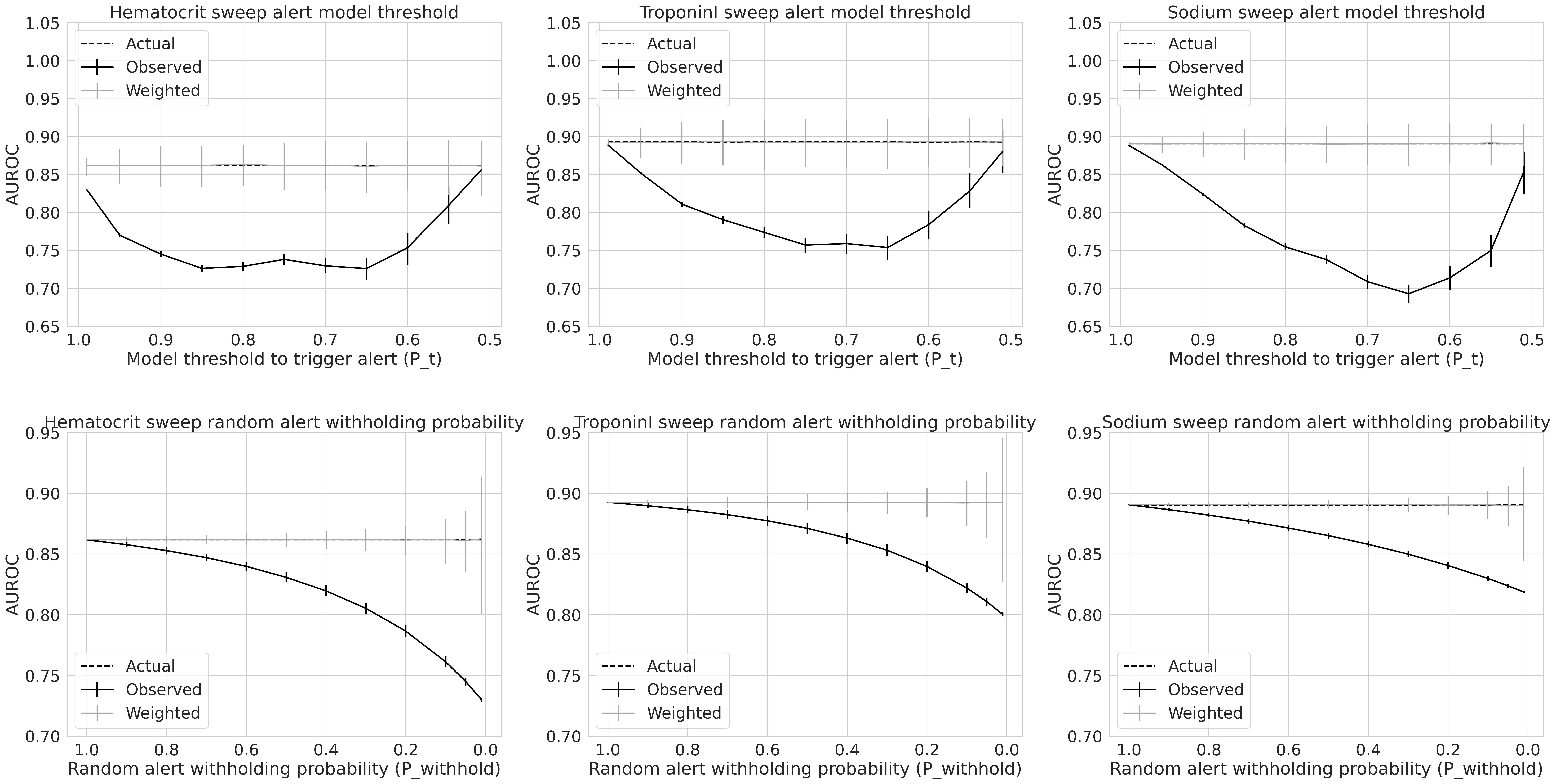}
\caption{Actual, observed and weighted AUROC as a function deployment simulation parameters}
\label{fig3}
\end{figure}

We then tackle the real-world task of monitoring the performance of deployed ML models whose interaction with clinicians feed-back and induce label selection. We train three ML models using real-world electronic medical record data that at order time predict whether stand-alone laboratory tests will result outside their normal range. We hypothesize a deployment use case in which upon order entry alerts are triggered in the electronic medical record suggesting not to order the tests. Alerts are triggered only if the probability estimate of the model is above threshold $p_t$ or below $1 - p_t$.  We then propose a slight adjustment to this production pipeline to enable consistent estimates of performance on the full deployment population. We propose injecting randomization into the alert triggering mechanism, such that predicted probabilities above $p_t$ or below $1 - p_t$  only trigger an alert if the result of a single sample from a Bernoulli distribution parameterized with $p=p_{withhold}$ is zero. By injecting randomization, and treating overridden alerts as missing, we ensure the selection mechanism is at random: only dependent on model predictions (a function of observed features) and a random number generator.  Further, we know the selection probability for each alert triggering example because we explicitly define it: $p_{withhold}$.  In deployment, labels of examples easiest to predict (furthest from the decision boundary) are more likely to go missing, resembling \textbf{Scenario 2} (select hard) from our simulations.  Our simulations reveal how naive model performance monitoring using the observed data leads to a large (up to 20\%) reduction in perceived discrimination acuity — perhaps large enough to cause an uninformed data science team to wrongfully scrap the project.  ML practitioners equipped with such a workflow can reliably monitor model performance even when deployment itself feeds back to induce label selection. 

\section*{Conclusion}
We simulate five causally distinct selection scenarios that affect the estimation of common binary ML performance metrics.  When selection is at random, naive estimates of discrimination acuity may be biased while calibration estimates remain intact.  When selection is not at random, naive estimates of calibration can be biased while discrimination metrics that condition on class labels (sensitivity, specificity, AUROC) remain intact.  We study the real-world use case of monitoring deployed ML models under feed-back mechanisms that induce label selection.  Naive estimates of performance lead to substantial (up to 20\%) reduction in perceived AUROC. We propose a deployment workflow that couples randomization and weighted estimates of performance and find that it consistently recovers true performance on the deployment population. 

\paragraph{Acknowledgements}
This research used data provided by STARR, STAnford medicine Research data Repository,” a clinical data warehouse containing de-identified Epic data from Stanford Health Care (SHC), the University Healthcare Alliance (UHA) and Packard Children’s Health Alliance (PCHA) clinics and other auxiliary data from Hospital applications such as radiology PACS. The STARR platform is developed and operated by the Stanford Medicine Research IT team and is made possible by Stanford School of Medicine Research Office.  The content is solely the responsibility of the authors and does not necessarily represent the official views of the NIH or Stanford Health Care.

\small
\bibliographystyle{vancouver}
\bibliography{main}

\end{document}